\documentclass[10pt,twocolumn,letterpaper]{article}
\usepackage{wacv}
\usepackage{times}
\usepackage{epsfig}
\usepackage{graphicx}
\usepackage{amsmath}
\usepackage{amssymb}
\usepackage{algorithm}
\usepackage{algorithmic}
\usepackage{adjustbox}

\wacvfinalcopy 

\author{Vishal Kaushal \\
IIT Bombay\\
{\tt\small vkaushal@cse.ittb.ac.in}
\and
Rishabh Iyer \\
Microsoft Corporation\\
{\tt\small rishi@microsoft.com}
\and
Khoshrav Doctor \\
University of Massachusetts, Amherst\\
{\tt\small kdoctor@cs.umass.edu}
\and
Anurag Sahoo \\
AitoeLabs\\
{\tt\small anurag@aitoelabs.com}
\and
Pratik Dubal \\
AitoeLabs\\
{\tt\small pratik@aitoelabs.com}
\and
Suraj Kothawade 
\\ IIT Bombay
\\{\tt\small surajkothawade@cse.iitb.ac.in}
\and
Rohan Mahadev \\
AitoeLabs\\
{\tt\small rohan@aitoelabs.com}
\and
Kunal Dargan \\
AitoeLabs\\
{\tt\small kunal@aitoelabs.com}
\and
Ganesh Ramakrishnan \\
IIT Bombay\\
{\tt\small ganesh@cse.ittb.ac.in}
}

\def\wacvPaperID{449} 

\ifwacvfinal\fi
\begin{document}

\title{Demystifying Multi-Faceted Video Summarization: Tradeoff Between Diversity, Representation, Coverage and Importance}

\maketitle

\ifwacvfinal\thispagestyle{empty}\fi

\begin{abstract}
This paper addresses automatic summarization of videos in a unified manner. In particular, we propose a framework for multi-faceted summarization for extractive, query base and entity summarization (summarization at the level of entities like objects, scenes, humans and faces in the video). We investigate several summarization models which capture notions of diversity, coverage, representation and importance, and argue the utility of these different models depending on the application. While most of the prior work on submodular summarization approaches has focused on combining several models and learning weighted mixtures, we focus on the explainability of different models and featurizations, and how they apply to different domains. We also provide implementation details on summarization systems and the different modalities involved. We hope that the study from this paper will give insights into practitioners to appropriately choose the right summarization models for the problems at hand.\looseness-1
\end{abstract}

\section{Introduction}
Visual Data in the form of images, videos and live streams have been growing at an unprecedented rate in the last few years. While this massive data is a blessing to data science by helping improve predictive accuracy, it is also a curse since humans are unable to consume this large amount of data. Moreover, today, machine generated videos (via Drones, Dash-cams, Body-cams, Security cameras, Go-pro etc.) are being generated at a rate higher than what we as humans can process. Moreover, majority of this data is plagued with redundancy. Given this data explosion, machine learning techniques which automatically understand, organize and categorize this data are of utmost importance. Video summarization attempts to provide a highlight of the most critical and important events in the video, giving the viewer a quick glimpse of the entire video so they can decide which parts of the video is important.

What comprises of the most critical aspect of a video depends largely on the domain. What is important in a surveillance video is very different from the highlights of a soccer game. This work attempts to provide a better understanding of different summarization models in different domains. We try to make a case that the choice of the summarization model really depends on the application and domain at hand. This paper investigates several choices of summarization models -- models which capture diversity, representation, importance or relevance, and coverage. We quantitatively and qualitatively study this pattern in many different domains, including surveillance footages, dashcam, bodycams and gopro footages, movies and TV shows and sports events like soccer. We argue how different characteristics are important for these domains, and through extensive experimentation establish the benefit of using the corresponding summarization models. For example, we show that for surveillance footages, diversity is more important compared to representation or coverage, while in a movie, representation and coverage form a better fit compared to diversity. Similarly, in a sports event like soccer, importance and relevance signals are important aspects of summarization. This paper also analyzes several choices of feature representations and concepts, including faces, scenes, humans, color information, objects etc. 

We study three variants of summarization: one is extractive summarization, the second is query focused summarization and the third is Entity based summarization (which we also call Concept based summarization). Entity based summarization focuses on entities, like objects, scenes, humans, faces to provide a representative yet diverse subset of these entities. This answers questions like who are the different people or what are the diverse objects and scenes in the video. Finally, we discuss several implementational details on how to create a video summarization system, including the preprocessing of features, different segmentations of shots and tricks for speeding up the optimization for near real time response times. 

\subsection{Existing Work}
Several papers in the past have investigated the problems of video and image collection summarization. Video Summarization techniques differ in the way they generate the output summary. Some of these~\cite{wolf1996key, lee2012discovering} extract a set of keyframes from the video, while others focus on extracting video summaries or skims from the long video~\cite{gygli2015video, zhang2016summary}. Other forms of video summarization include creating GIF summaries from videos~\cite{gygli2016video2gif}, Montages~\cite{sun2014salient}, Visual Storyboards from videos~\cite{goldman2006schematic}, video synopses~\cite{pritch2008nonchronological} and time lapses and hyperlapse summaries~\cite{kopf2014first}. Similarly, image collection summarization involves choosing a subset of representative  images from the collection~\cite{tschiatschek2014learning}. Another line of approach, which is similar to what we call Entity based Summarization, was proposed in \cite{meng2016keyframes}, wherein the authors select representative summaries of all objects in a video. They do this by modeling the problem as that of sparse dictionary selection. Most video summarization techniques can be categorized into methods trying to model one of three properties of summaries (i) interestingness (how good is a given snippet as a summary), (ii) representativeness (how well the summary represents the entire video or image collection), and (iii) diversity (how non-redundant and diverse is the summary). 

Examples of methods which model interestingness of snippets include~\cite{wolf1996key} that find summary snippets through motion analysis and optical flow, \cite{lee2012discovering} which uses humans and objects to determine interesting snippets and finally, \cite{gygli2014creating} which models interestingness through a super-frame segmentation. \cite{chu2015video} summarizes multiple videos collectively by looking at inter-video-frame similarity and posing a maximal bi-clique finding algorithm for finding summaries. Methods which only model the quality of the snippets, or equivalently the interestingness of the summaries and do not model the diversity often achieve redundant frames and snippets within their summary. 

Hence a lot of recent work has focused on diversity models for video and image collection summarization. \cite{simon2007scene} used the Facility Location function with a diversity penalty for image collection summarization, while \cite{sinha2011extractive} defined a coverage function and a disparity function as a diversity model. \cite{lu2013story} attempted to find the candidate chain of sub shots that has the maximum score composed of measures of story progress between sub shots, importance of individual sub shots and diversity among sub shot transitions.   \cite{tschiatschek2014learning} was among the first to use a mixture of submodular functions learnt via human image summaries for this problem. For video summarization, \cite{li2010multi} proposed the Maximum Marginal Relevance (MMR) as a diversity model, while \cite{zhang2016summary, gong2014diverse} used a Determinantal Point Process based approach for selecting diverse summaries. \cite{zhao2014quasi} proposed an approach for video summarization based on dictionary based sparse coding, and \cite{gygli2015video} proposed using mixtures of submodular functions and supervised learning of these mixtures via max-margin training, an approach used for several other tasks including document summarization~\cite{lin2012learning} and image collection summarization~\cite{tschiatschek2014learning}. 

\subsection{Our Contributions}
The goal of this work is not to achieve the best results on Video and Image summarization tasks and datasets like TVSum~\cite{tvsumm} and Summe~\cite{GygliECCV14}. Rather, we attempt to provide insights into what it takes to build a real world video summarization system. In particular, we try to understand the role of different submodular functions in different domains, and how to implement a video summarization system in practice. 

As observed in prior work~\cite{gygli2015video,lin2012learning,tschiatschek2014learning} several models for diversity, representation, coverage and uniformity can be unified within the class of Submodular Optimization. We build upon this work as follows.

\begin{enumerate}
\item This paper studies the role and characteristics of different summarization models. What constitutes a good summary depends on the particular domain at hand.
\item We investigate several diversity, coverage and representation models, and demonstrate how different models are applicable in different kinds of video summarization tasks. 
\item We validate our claims by empirically showing the behavior of these functions on different kinds of videos, and quantitatively prove this on several videos in each domain. For example, we show that \emph{Diversity} models focus on getting outliers in the video, which is important in domains like surveillance. On the other hand, \emph{Representation} models capture the centroids and important scenes, which is useful in Movies. We also argue how coverage functions focus on achieving a good coverage of concepts. Similarly, we show that in domains like soccer, \emph{Importance} or \emph{Relevance} plays the most important role in the summary. 
\item We also discuss the computational scalability of the optimization algorithms, and point out some computational tricks including lazy evaluations and memoization, which enable optimized implementations for various submodular functions. As a result, we show that once the important visual features have been extracted (via a pre-processing step), we can obtain the summary subset of the video (or frames) in a few seconds. This allows the user to interactively obtain summaries of various lengths, types and queries in real time. We empirically demonstrate the benefit of memoization and lazy greedy implementations for various video summarization problems.
\end{enumerate}

Most past work on Video and Image collection summarization, either use a subset of hand-tuned submodular functions~\cite{simon2007scene,lin2012learning,li2010multi} or a learnt mixture of submodular functions~\cite{gygli2015video,tschiatschek2014learning,lin2012learning}. This work addresses the orthagonal aspect how do different subclasses of submodular functions model summarization and their performance in different video domains. We believe the insights gathered from this work, will help practitioners in choosing appropriate models for several real world video and image summarization tasks. 

\section{Background and Main Ideas}
This section describes the building blocks of our framework, namely the Submodular Summarization Framework and the basics of Convolutional Neural Networks for Image recognitions (to extract all the objects, scenes, faces, humans etc.)
\subsection{Submodular Summarization Framework}
We assume we are given a set $V = \{1, 2, 3, \cdots, n\}$ of items which we also call the \emph{Ground Set}. Also define a utility function $f:2^V \rightarrow \mathbf{R}$, which measures how good of a summary a set $X \subseteq V$ is. Let $c :2^V \rightarrow \mathbf{R}$ be a cost function, which describes the cost of the set (for example, the size of the subset). The goal is then to have a summary set $X$ which maximizes $f$ while simultaneously minimizes the cost function $c$. In this paper, we study a special class of set functions called \emph{Submodular Functions}. Given two subsets $X \subseteq Y \subseteq V$, a set function $f$ is submodular, if $f(X \cup j) - f(X) \geq f(Y \cup j) - f(j)$, for $j \notin Y$. This is also called the diminishing returns property.  Several Diversity and Coverage functions are submodular, since they satisfy this diminishing returns property. We also call a function \emph{Monotone Submodular} if $f(X) \leq f(Y)$, if $X \subseteq Y \subseteq V$. The ground-set $V$ and the items $\{1, 2, \cdots, n\}$ depend on the choice of the task at hand. We now define a few relevant optimization problems which shall come up in our problem formulations:
\begin{align}
\mbox{Problem 1:} \max_{X \subseteq V, s(X) \leq b} f(X)
\end{align}
Problem 1 is knapsack constrained submodular maximization~\cite{sviridenko2004note}. The goal here is to find a summary with a fixed cost, and $s_1, s_2, \cdots, s_n$ denotes the cost of each element in the ground-set. A special case is cardinality constrained submodular maximization, when the individual costs are $1$~\cite{nemhauser1978analysis}. This a natural model for extracting fixed length summary videos (or a fixed number of keyframes). 

\begin{align}
\mbox{Problem 2: } \min_{f(X) \geq c} s(X)
\end{align}
This problem is called the \emph{Submodular Cover Problem}~\cite{wolsey1982analysis,iyer2013submodular}. $s(X)$ is the modular cost function, and  $c$ is the coverage constraint. The goal here is to find a minimum cost subset $X$ such that the submodular coverage or representation function covers \emph{information} from the ground set. A special case of this is the set cover problem. Moreover, Problem 2 can be seen as a Dual version of Problem 1~\cite{iyer2013submodular}.

Submodular Functions have been used for several summarization tasks including Image summarization~\cite{tschiatschek2014learning}, video summarization~\cite{gygli2015video}, document summarization~\cite{lin2012learning}, training data summarization and active learning~\cite{wei2015submodularity} etc. Using a
greedy algorithm to optimize a submodular function (for selecting a subset) gives a lower-bound performance guarantee of around 63\% of optimal and in practice these greedy solutions are often within 90\% of optimal ~\cite{krause2008optimizing}. This makes it advantageous to formulate (or approximate) the objective function for data selection as a submodular function. 

\subsection{CNNs for Image Feature Extraction}
Convolutional Neural Networks are critical to feature extraction in our summarization framework. We pre-process the video to extract key visual features including objects, scenes, faces, humans, etc. Convolutional Neural Networks have recently provided state of the art results for several recognition tasks including object recognition~\cite{krizhevsky2012imagenet,szegedy2015going,he2016deep}, Scene recognition~\cite{zhou2014learning}, Face Recognition~\cite{parkhi2015deep} and Object Detection and Localization~\cite{redmon2016you}. We next describe the end to end system in detail.

\section{Method}
The input to our system is a video. Our system then extracts all important features from the video and generates an analysis database. The user can then interact with the system in several ways. User can generate a video summary of a given length, or extract a set of key frames or a montage describing the video. Similarly the user can search for a query and extract video snippets of frames which are relevant to the query. Finally the user can also view a summary of all objects, scenes, humans and faces in the video along with their statistics. All these interactions are enabled on the fly (in a few seconds). The user can also define the summarization model of their choice. We investigate and compare  different submodular models, and argue the utility of different models based on the use case. 
\subsection{Problem Formulation for the Multi-Faceted Visual Summarization}
We now formulate problem statements across the different summarization views. Extractive summarization considers the entire video. We can generate a summary either in terms of key frames (represent the video as a set of frames sampled at a frame-rate), or video snippets. In either case, we extract a ground set $V$, with each individual element either being a key frame or a video snippet. We then solve Problems 1 or 2 depending on the use case. Problem 1 is the right formulation if we are interested in obtaining a summary of a fixed budget. Problem 2 is useful if we don't care about the size of the video, but we are interested in the summary capturing all the \emph{information} of the video. In the case of query based summarization, we first extract the set of frames or snippets relevant to that query $q$. Denote this by $V_q$.  We then solve the submodular optimization problem on $V_q$. Finally, in the case of entity based summarization, we extract all the entities in the video, and denote the set of entities as $V_e$. $V_e$ represents, for example, all the faces of people in the video. We can then run our summarization with $V_e$ as the groundset. 

In the case of Extractive or Query based summarization, the ground truth elements can be either frames of video snippets. Our video snippets can be either fixed length snippets or Shots, obtained from a shot detector. If the snippets are fixed length snippets (say, 2 or 3 seconds), we can use the cardinality constrained submodular maximization. If the snippets are shots from the video, the length of each shot can differ, and we have the more general knapsack constrained setting. While our system can handle each of these modes, we focus on the key-frame based method for our experiments, since we are interested in proving the utility of different summarization models. The insights will carry over to the other modes as well.

\begin{figure*}
\begin{center}
\includegraphics[width = 0.3\textwidth]{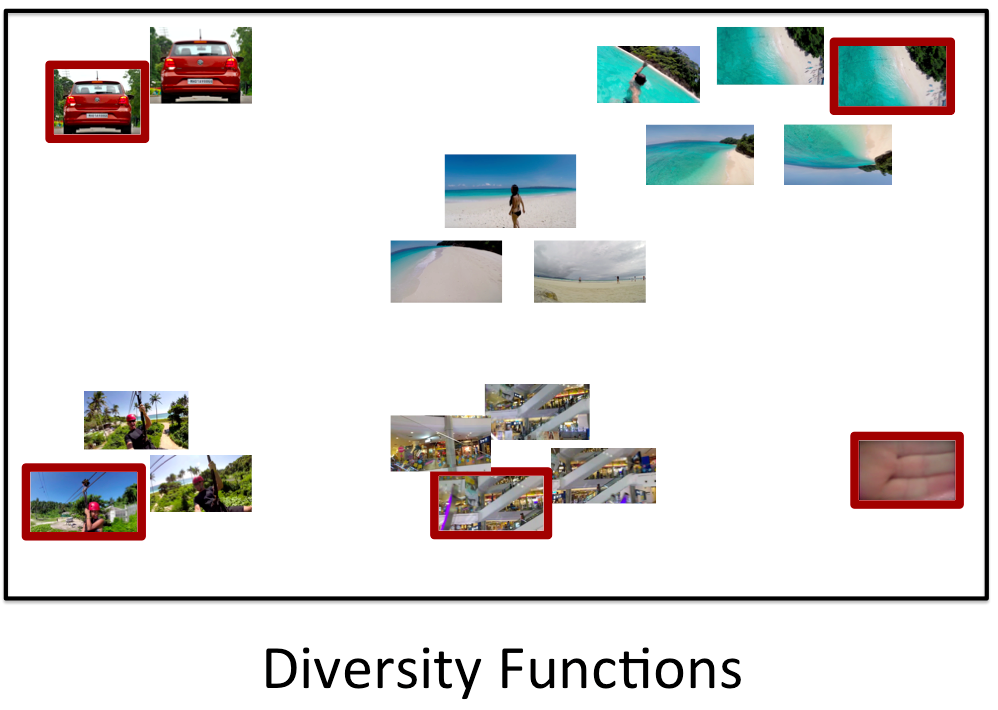}
\includegraphics[width = 0.3\textwidth]{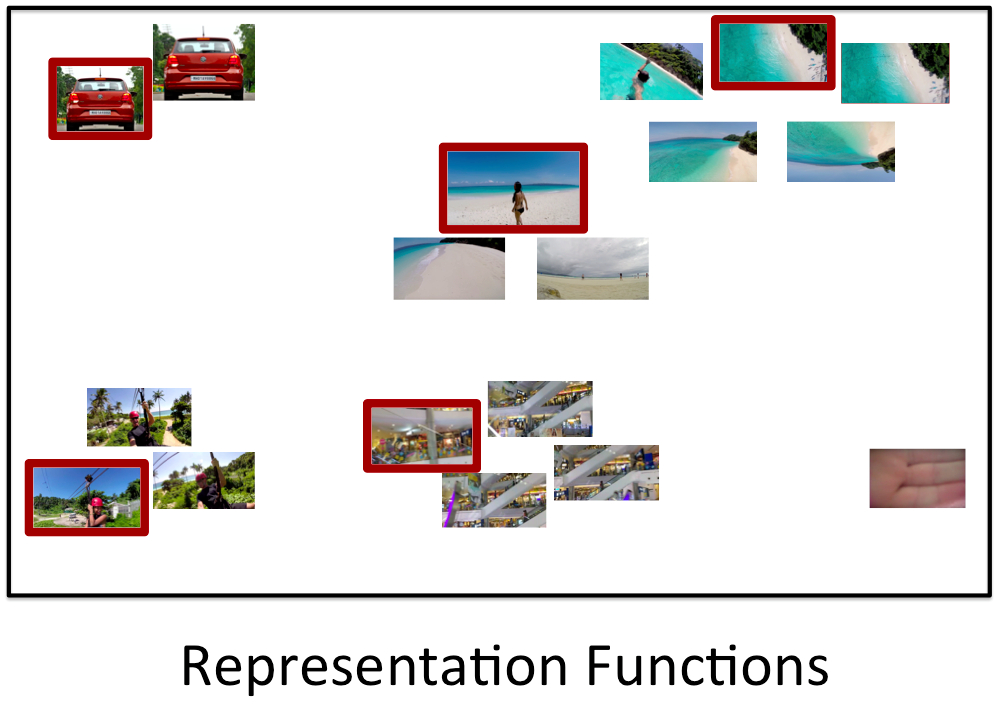}
\includegraphics[width = 0.3\textwidth]{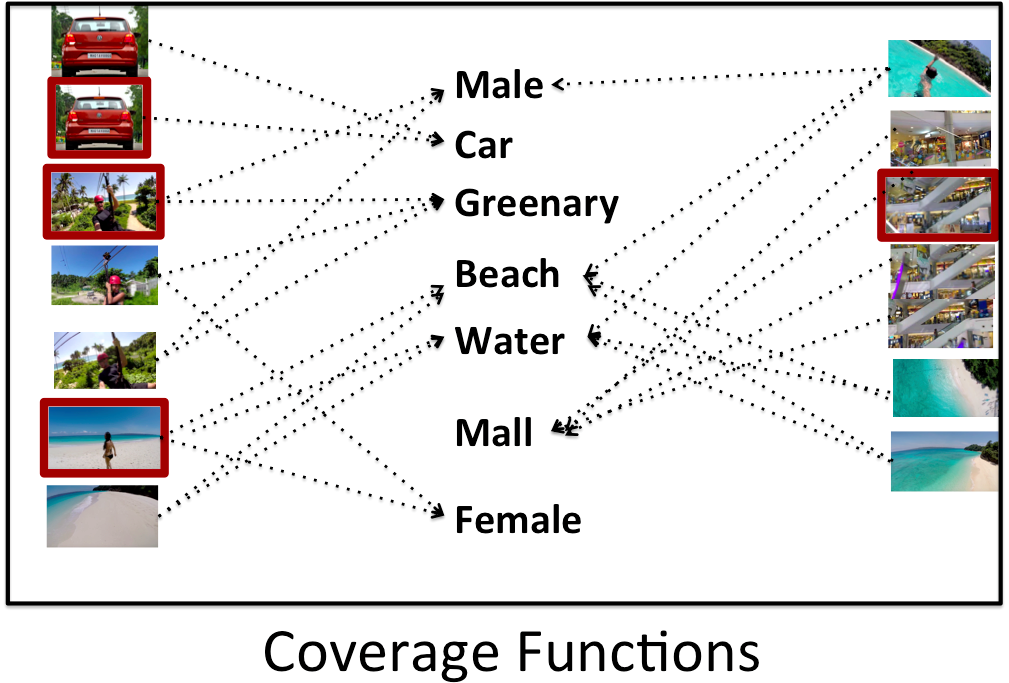}
\end{center}
\caption{Illustration of the Difference between Diversity Functions, Coverage Functions and Representation Functions}
\label{divrepcov}
\end{figure*}

\subsection{Submodular Functions as Summarization Models}
This section describes the Submodular Functions used in our system. We divide these into Coverage Functions, Representation Functions and Diversity Functions.

\subsubsection{Modeling Coverage}
This class of functions model notions of coverage, i.e. try to find a subset of the ground set $X$ which covers a set of \emph{concepts}. Below are instantiations of this.

\noindent \textbf{Set Cover Function:} Denote $V$ as the ground set and let $X \subseteq V$ be a subset (of snippets or frames). Further $\mathcal U$ denotes a set of concepts, which could represent, for example, scenes or objects. Each frame (or snippet) $i \in X$ contains a subset $U_i \in \mathcal U$ set of concepts (for example, an image covers a table, chair and person). The set cover function then is 
\begin{align}
f(X) = w(\cup_{i \in X} U_i), 
\end{align}
where $w_u$ denotes the weight of concept $u$. 

\noindent \textbf{Probabilistic Set Cover: } This is a generalization of the set cover function, to include probabilities $p_{i u_i}$ for each object $u_i$ in Image $i \in X$. For example, our convolutional neural network might output a confidence of object $u_i$ in Image $i$, and we can use that in our function. The probabilistic coverage function is defined as, 
\begin{align}
f(X) = \sum_{i \in \mathcal U} w_i[1 - \prod_{i \in X}(1 - p_{ij})].
\end{align}
The set cover function is a special case of this if $p_{ij} = 1$ if Object $j$ belongs to Image $i$ (i.e. we use the hard labels instead of probabilities).

\noindent \textbf{Feature Based Functions: } Finally we investigate the class of Feature Based functions. Here, we denote an Image $i$ via a feature representation $q_i$. This could be, for example, the features extracted from the second last layer of a ConvNet. Denote $F$ as the set of features. The feature based function is defined as,
\begin{align}
f(X) = \sum_{i \in F} \psi(q_i(X))
\end{align}
where $q_i(X) = \sum_{j \in X} q_{ij}$, and $q_{ij}$ is the value of feature $i$ in Image $j$. $\psi$ is a concave function.  Examples of $\psi$ are square-root, Log and Inverse Function etc. 

\subsubsection{Modeling Representation}
Representation based functions attempt to directly model representation, in that they try to find a representative subset of items, akin to centroids and mediods in clustering.

\noindent \textbf{Facility Location Function: } The Facility Location function is closely related to k-mediod clustering. Denote $s_{ij}$ as the similarity between images $i$ and $j$. We can then define $f(X) = \sum_{i \in V} \max_{j \in X} s_{ij}$. For each image $i$, we compute the representative from $X$ which is closest to $i$ and add the similarities for all images. Note that this function, requires computing a $O(n^2)$ similarity function. However, as shown in~\cite{wei2014fast}, we can approximate this with a nearest neighbor graph, which will require much smaller space requirement, and also can run much faster for large ground set sizes.

\noindent \textbf{Saturated Coverage Function: } The saturated coverage function~\cite{lin2011class} is defined as $f(X) = \min\{\sum_{i \in X} s_{ij}, \alpha \sum_{i \in V} s_{ij}\}$. This function is similar to Facility Location and attempts to model representation. This is also a Kernel based function and requires computing the similarity matrix.

\noindent \textbf{Graph Cut Functions: } We define the graph cut family of functions as $f(X) = \lambda \sum_{i \in V} \sum_{j \in X} s_{ij} - \sum_{i, j \in X} s_{ij}$. This function is similar to the Facility Location and Saturated Coverage in terms of its modeling behaviour. 

\subsubsection{Modeling Diversity}
The third class of Functions are Diversity based ones, which attempt to obtain a diverse set of key points. 

\noindent \textbf{Dispersion (Disparity) Functions: } Denote $d_{ij}$ as a distance measure between Images $i$ and $j$. Define a set function $f(X) = \min_{i, j \in X} d_{ij}$. This function is not submodular, but can be efficiently optimized via a greedy algorithm~\cite{dasgupta2013summarization}. It is easy to see that maximizing this function involves obtaining a subset with maximal minimum pairwise distance, thereby ensuring a diverse subset of snippets or keyframes. Similar to the Minimum Disparity, we can define two more variants. One is Disparity Sum, which can be defined as $f(X) = \sum_{i, j \in X} d_{ij}$. This is a supermodular function. Another model is, what we call, Disparity Min-Sum which is a combination of the two forms of models. Define this as $f(X) = \sum_{i \in X} \min_{j \in X} d_{ij}$. This function is submodular~\cite{chakraborty2015adaptive}. 

\noindent \textbf{Determinantal Point Processes: } Another class of Functions are Determinantal Point Processes, defined as $p(X) = \mbox{Det}(S_X)$ where $S$ is a similarity kernel matrix, and $S_X$ denotes the rows and columns instantiated with elements in $X$. It turns out that $f(X) = \log p(X)$ is submodular, and hence can be efficiently optimized via the Greedy algorithm. Unlike the other choices of submodular functions investigated so far, this requires computing the determinant and is $O(n^3)$ where $n$ is the size of the ground set. This function is not computationally feasible and hence we do not use it in our system since we require near real time results in summarization.

\begin{table*}
\begin{center}
 \begin{tabular}{|| c | c | c |  c | c ||} 
 \hline
 Name & $f(X)$ & $p_f(X)$ & $T^o_f$ & $T^p_f$\\ [0.5ex] 
 \hline
 Facility Location & $\sum_{i \in V} \max_{k \in X} s_{ik}$ & $[\max_{k \in X} s_{ik}, i \in V]$ & $O(n^2)$ & $O(n)$\\ 
 \hline
 Saturated Coverage & $\sum_{i \in V} \min\{\sum_{j \in X} s_{ij}, \alpha_i\}$ & $[\sum_{j \in X} s_{ij}, i \in V]$ & $O(n^2)$ & $O(n)$\\ 
\hline
 Graph Cut & $\lambda \sum_{i \in V}\sum_{j \in X} s_{ij} - \sum_{i, j \in X} s_{ij}$ & $[\sum_{j \in X} s_{ij}, i \in V]$ & $O(n^2)$ & $O(n)$\\ 
\hline
 Feature Based & $\sum_{i \in \mathcal F} \psi(w_i(X))$ & $[w_i(X), i \in \mathcal F]$ & $O(n|\mathcal F|)$ & $O(|\mathcal F|)$ \\
 \hline
 Set Cover & $w(\cup_{i \in X} U_i)$ & $\cup_{i \in X} U_i$ & $O(n|U|$ & $|U|$\\
 \hline
 Prob. Set Cover & $\sum_{i \in \mathcal U} w_i[1 - \prod_{k \in X}(1 - p_{ik})]$ & $[\prod_{k \in X} (1 - p_{ik}), i \in \mathcal U]$ & $O(n|\mathcal U|)$ & $O(|\mathcal U|)$ \\ [1ex] 
 \hline
 DPP & $\log\det(S_X))$ & SVD($S_X$) & $O(|X|^3)$ & $O(|X|^2)$\\
 \hline
 Dispersion Min & $\min_{k,l  \in X, k \neq l} d_{kl}$ & $\min_{k, l \in X, k \neq l} d_{kl}$ & $O(|X|^2)$ & $O(|X|)$\\
 \hline
 Dispersion Sum & $\sum_{k,l  \in X} d_{kl}$ & $[\sum_{k \in X} d_{kl}, l \in X]$ & $O(|X|^2)$ & $O(|X|)$\\
  \hline
 Dispersion Min-Sum& $\sum_{k \in X} \min_{l \in X} d_{kl}$ & $[\min_{k \in X} d_{kl}, l \in X]$ & $O(|X|^2)$ & $O(|X|)$\\
 \hline
\end{tabular}
\caption{List of Submodular Functions used, with the precompute statistics $p_f(X)$, gain evaluated using the precomputed statistics $p_f(X)$ and finally $T^f_o$ as the cost of evaluation the function without memoization and $T^f_p$ as the cost with memoization. It is easy to see that memoization saves an order of magnitude in computation.}
\end{center}
\end{table*}

\subsubsection{Modeling Importance and Relevance}
To model Importance or Relevance, we use Modular terms~\cite{potapov2014category}. Given a specific task, we train a supervised model to predict the important frames in that video (for example, a \emph{goal} might be considered important in a soccer video). Given this learnt model, we can predict the score of each frame, and rank the scores. This is exactly equivalent to optimize the modular function defined with these scores.

\subsubsection{Understanding Diversity, Representation and Coverage}
Figure~\ref{divrepcov} demonstrates the intuition of using diversity, representation and coverage functions. Diversity based functions attempt to find the most different set of images. The leftmost figure in Fig.~\ref{divrepcov} demonstrates this. It is easy to see that the five most diverse images are picked up by the diversity function (Disparity Min), and moreover, the summary also contains the image with a hand covering the camera (the image on the right hand side bottom), which is an outlier. The middle figure demonstrates the summary obtained via a representation function (like Facility Location). The summary does not include outliers, but rather contains one representative image from each cluster. The diversity function on the other hand, does not try to achieve representation from every cluster. The third figure demonstrates coverage functions. The summary obtained via a coverage function (like Set Cover or Feature based function), covers all the concepts contained in the images (Male, Car, Greenery, Beach etc.). 

\subsection{Instantiations of the Submodular Functions}
Having discussed the choices of the submodular functions and features, we go over the specific instantiations of submodular functions considered in our system. First consider Extractive and Query Based Summarization. For the Facility Location function and the disparity min function, we define the similarity kernel as:
\begin{align*}
s_{ij} = \langle F_s^i, F_s^j \rangle + \langle F_o^i, F_o^j \rangle + \mbox{corr}(H^i, H^j)
\end{align*}
where $F_s$ represent normalized Deep Scene Features extracted using GoogleNet on Places205~\cite{zhou2014learning}, $F_o$ represents normalized Deep Object features using GoogleNet on ImageNet~\cite{szegedy2015going}, $H$ represents the normalized color histogram features. Since the disparity min function uses a distance function, we use $d_{ij} = 1 - s_{ij}$. For Feature based functions, the feature-set $\mathcal F$ is a concatenation of the scene features $F_s$ and object features $F_o$. In order to define the Set Cover function, we define $U_i$ as the Scene and YOLO object labels corresponding to the Image. Recall that the labels for scenes and objects were chosen based on a pre-defined threshold (i.e. select scene and objects labels if the probability for the label is greater than a threshold). The Probabilistic Set Cover function is defined via a concatenation of the probabilities from the scene and object models. Query based summarization for keyframes is identical to extractive summarization, except that we first get a groundset $V_q$ which is related to the query. The queries, are either objects, scenes, faces/humans with age and gender, text in the video, as well as meta data like subtitles etc. 

Finally, for entity or concept based summarization, we extract the entities from the videos. Entities we consider are objects, faces etc. For faces, we use the VGG Face model from~\cite{parkhi2015deep}, pretrained on Celeb Face data for Face recognition. The objects are localized using YOLO~\cite{redmon2016you}. We extract features from GoogLeNet~\cite{krizhevsky2012imagenet,szegedy2015going}, along with color histogram~\cite{swain1991color}. The similarity kernel we use here is $s_{ij} = \langle F_o^i, F_o^j \rangle + \mbox{corr}(H^i, H^j)$. 


Next we discuss the choice of the submodular functions. 
Facility Location, Disparity Min/Sum, Graph Cut, Saturated Coverage, and DPPs are instantiated using Similarity Kernels discussed above. Feature Based functions are defined directly via features, and we use the deep features as described above. In the case of the Set cover and probabilistic set cover functions, we use the labels and probabilities respectively from the deep models as the concepts. 


\subsection{Optimization Algorithms}
The previous sections describe the models used in our system. We now investigate optimization algorithms which solve Problems 1 and 2. Variants of a greedy algorithm provide near optimal solutions with approximation guarantees for Problems 1-3~\cite{wolsey1982analysis,nemhauser1978analysis,sviridenko2004note}. 


\noindent \textbf{Budget Constrained Submodular Maximization: } For the budget constrained version (Problem 1), the greedy algorithm is a slight variant, where at every iteration, we sequentially update $X^{t+1} = X^t \cup \mbox{argmax}_{j \in V \backslash X^t} \frac{f(j | X^t)}{c(j)}$. This algorithm has near optimal guarantees~\cite{sviridenko2004note}. 

\noindent \textbf{Submodular Cover Problem: } For the Submodular Cover Problem (Problem 2), we again resort to a greedy procedure~\cite{wolsey1982analysis} which is near optimal. In this case, the update is similar to that of problem 1, i.e. choose $X^{t+1} = X^t \cup \mbox{argmax}_{j \in V \backslash X^t} f(j | X^t)$. We stop as soon as $f(X^t) = f(V)$, or in other words, we achieve a set which covers all the concepts.

\noindent \textbf{Lazy Greedy Implementations: } Each of the greedy algorithms above admit lazy versions which run much faster than the worst case complexity above~\cite{minoux1978accelerated}. The idea is that instead of recomputing $f(j | X^t), \forall j \notin ^t$, we maintain a priority queue of sorted gains $\rho(j), \forall j \in V$. Initially $\rho(j)$ is set to $f(j), \forall j \in V$. The algorithm selects an element $j \notin X^t$, if $\rho(j) \geq f(j | X^t)$, we add $j$ to $X^t$ (thanks to submodularity). If $\rho(j) \leq f(j | X^t)$, we update $\rho(j)$ to $f(j | X^t)$ and re-sort the priority queue. The complexity of this algorithm is roughly  $O(k n_R T_f)$, where $n_R$ is the average number of re-sorts in each iteration. Note that $n_R \leq n$, while in practice, it is a constant thus offering almost a factor $n$ speedup compared to the simple greedy algorithm.

\section{Implementational Tricks}
This section goes over implementation tricks via memoization. One of the parameters in the lazy greedy algorithms is $T_f$, which involves evaluating $f(X \cup j) - f(X)$. One option is to do a na\"{\i}ve implementation of computing $f(X \cup j)$ and then $f(X)$ and take the difference. However, due to the greedy nature of algorithms, we can use memoization and maintain a precompute statistics $p_f(X)$ at a set $X$, using which the gain can be evaluated much more efficiently. At every iteration, we evaluate $f(j | X)$ using $p_f(X)$, which we call $f(j | X, p_f)$.  We then update $p_f(X \cup j)$ after adding element $j$ to $X$. Table 1 provides the precompute statistics, as well as the computational gain for each choice of a submodular function $f$. Denote $T_f^o$ as the time taken to na\"{\i}vely compute $f(j | X) = f(X \cup j) - f(X)$. Denote $T_o^p$ as the time taken to evaluate this gain given the pre-compute statistics $p_X$. We see from Table 1, that evaluating the gains using memoization is often an order of magnitude faster. Moreover, notice that we also need to update the pre-compute statistics $p_X$ at every iteration. For the functions listed in Table 1, the cost of updating the pre-compute statistics is also $T_f^p$. Hence every iteration of the (lazy) greedy algorithm costs only $2T_f^p$ instead of $T_f^o$ which is an order of magnitude larger in every case. In our results section, we evaluate empirically the benefit of memoization in practice.

\begin{table}
    \centering
    \begin{tabular}{ |c|c|c|c|c|c|c|}
        \hline
         & \multicolumn{3}{c|}{\textbf{Memoization}} & \multicolumn{3}{c|}{\textbf{No Memoization}} \\
        \hline
        \textbf{Function} & \textbf{5\%} & \textbf{15\%} & \textbf{30\%} & \textbf{5\%} & \textbf{15\%} & \textbf{30\%}  \\
         \hline
        Fac Loc & 0.34 & 0.4 & 0.71 & 48 & 168 & 270  \\
         \hline
        Sat Cov & 0.36 & 0.64 & 0.92 & 55 & 177 & 301  \\
        \hline
         Gr Cut & 0.39 & 0.52 & 0.82 & 41 & 161 & 355  \\
        \hline
        Feat B & 0.16 & 0.21 & 0.32 & 9 & 16 & 21  \\
        \hline
        Set Cov & 0.21 & 0.31 & 0.41 & 5 & 16 & 31  \\
        \hline
        PSC & 0.11 & 0.37 & 0.42 & 7 & 19 & 35  \\
        \hline
        DPP & 32 & 107 & 411 & 171 & 1003 & 4908  \\
                \hline
        DM & 0.11 & 0.61 & 0.82 & 21 & 125 & 221  \\
                \hline
        DS & 0.21 & 0.63 & 0.89 & 41 & 134 & 246  \\
        \hline
    \end{tabular}
    \caption{Timing results in seconds for summarizing a two hour video for various submodular functions}
    \label{tab:my_label}
\end{table}

\begin{figure}
\includegraphics[width=0.5\textwidth, clip=true]{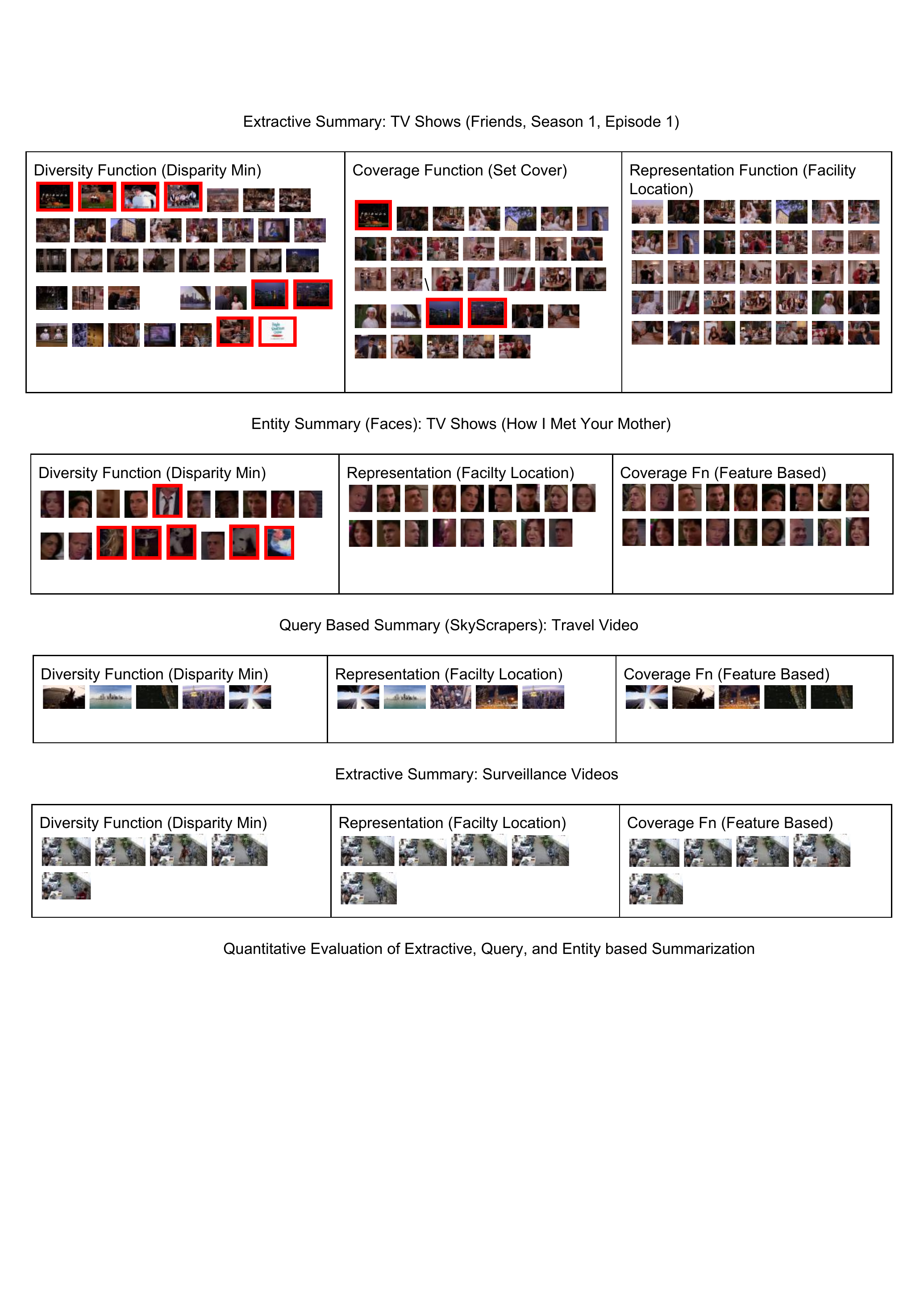}
\vspace{-3.3cm}
\caption{Illustration of the Results. The top figure shows the results from extractive summarization on TV shows, the second demonstrates entity summary on a TV show. The third figure shows the results of query based summarization on a query "SkyScraper" while the fourth one shows the results of extractive summarization on surveillance videos. In each case we compare Representation, Diversity and Coverage models. See the text for more details.}
\label{results}
\end{figure}

\begin{figure}
\centering{
\includegraphics[width=0.21\textwidth]{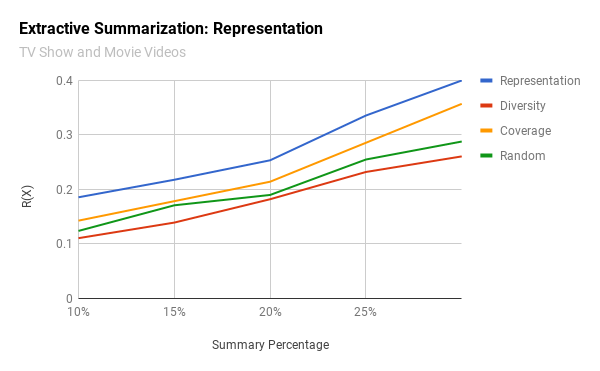}
~
\includegraphics[width=0.21\textwidth]{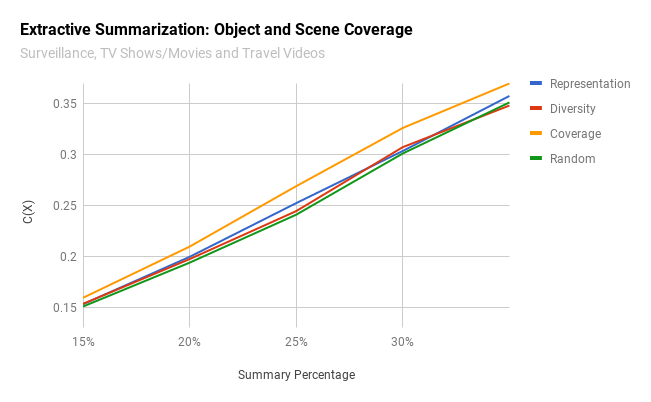}

\includegraphics[width=0.21\textwidth]{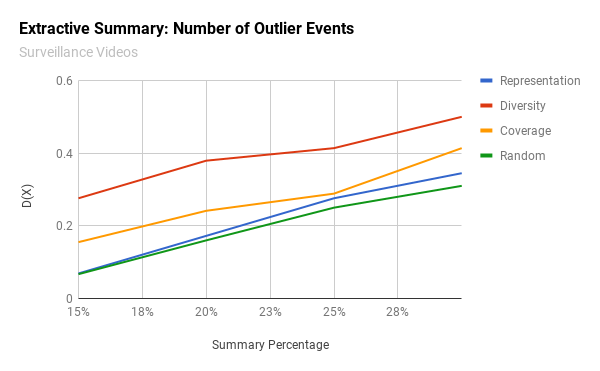}
~
\includegraphics[width=0.21\textwidth]{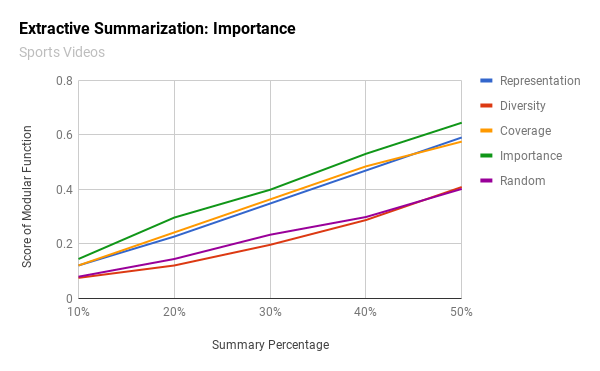} 

\includegraphics[width=0.21\textwidth]{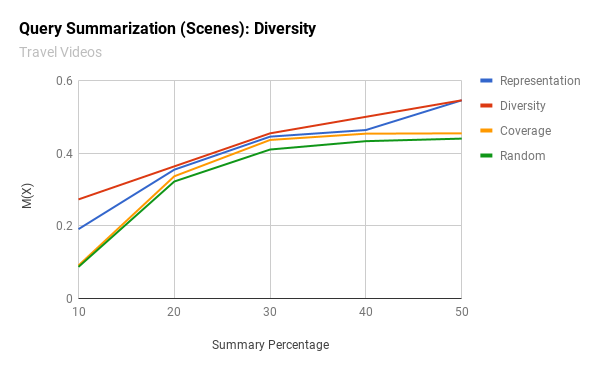}
~
\includegraphics[width=0.21\textwidth]{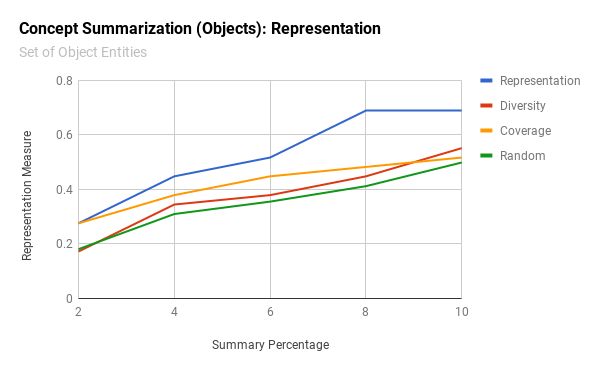}}
\caption{Comparison of Diversity, Coverage and Representation Models for various domains and scenarios. See the text for more details.}
\label{quantres}
\end{figure}

\begin{figure*}
\includegraphics[width=\textwidth,height=4cm]{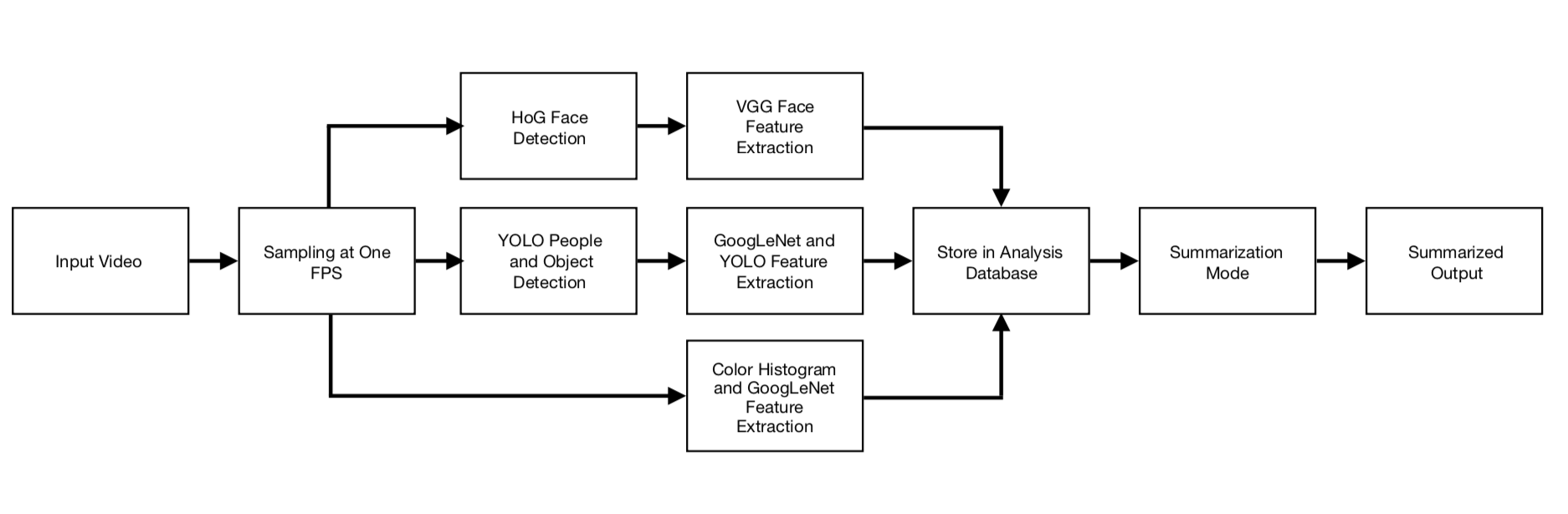}
\caption{End-to-End Processing and Summarization of a Video}
\label{systemFlowchart}
\end{figure*}
\section{Results}
Our system is implemented in C++. We use Caffe~\cite{jia2014caffe} and DarkNet~\cite{redmon2016you} for deep CNNs and OpenCV for other computer vision tasks. A graphical representation of our system is depicted in Figure~\ref{systemFlowchart}.


Figure~\ref{results} shows the results for extractive summarization as keyframes, extractive summarization on concepts or entities and query based summarization on keyframes. We compare the different summarization models under various scenarios and evaluation measures. Instead of comparing all the submodular functions described above, we consider representatives from each class of functions. We use Facility Location as a representative function, Disparity Min for Diversity and Set Cover as a choice for Coverage functions.

We next create a dataset of videos from different categories. We select 10 videos from the following categories: Movies/TV shows, Surveillance camera footage, Travel videos and sports videos like Soccer. In the following sections, we annotate various events of interest (ground-truth) from these videos to define various evaluation criteria. The annotation mechanism and evaluation criteria is described in each of the sections below. The goal of this is to demonstrate the role of various summarization models. 

\paragraph{Extractive Summarization: Representation} The top Figure in Fig.~\ref{results} demonstrates the results of extractive summarization on Movies and TV shows. Diversity Models tend to pick up outlier events, which in this case, include transition scenes and other outliers. In contrast, the Representation function (Facility Location) tends to pick the representative scenes. The coverage function does something in between. In the case of a TV show, representative shots are probably more important compared to the transition scenes. To quantify this, define an evaluation measure as follows. We divide a movie (TV Show) into a set of scenes $S_1, \cdots, S_k$ where each scene $S_i$ is a continuous shot of events. We do not include the outliers (we define outliers as shots less than a certain length -- for example transition scenes). Given a summary $X$, define $R(X) = \sum_{i = 1}^k \min(|X \cap S_i|, 1)/k$. A summary with a large value of $R(X)$ will not include the outliers and will pick only single representatives from each scene. We evaluate this on $10$ different TV show and movie videos. Figure~\ref{quantres} (top left) compares the representative, diversity, and coverage models and a random summary baseline. We see the representative model (Facility Location) tends to perform the best as expected, followed by the coverage model. The diversity model does poorly since it picks a lot of outliers. 

\paragraph{Extractive Summarization: Coverage} Next, we define an evaluation criteria capturing coverage. For each frame in the video (sampled at 1FPS), define a set of concepts covered $\mathcal U$. Denote $\mathcal U(X)$ as the set of concepts covered by a set $X$. For each frame of the video, we hand pick a set of concepts (scenes and objects contained in the video). Define the coverage objective as $C(X) = \mathcal U(X)/\mathcal U(V)$. Figure~\ref{quantres} demonstrates the coverage objective for the different models. We obtain this by creating a set of 10 labeled videos of different categories (surveillance, TV shows/movies, and travel videos). As expected, the coverage function (set cover) achieves superior results compared to the other models. 

\paragraph{Extractive Summarization: Outliers and Diversity} In the above paragraphs, we define two complementary evaluation criteria, one which captures representation and another which measures coverage. We argue how, for example, representation is important in Movies and TV shows. We now demonstrate how the diversity models tend to select outliers and anomalies. To demonstrate this, we select a set of surveillance videos. Most of our videos have repetitive events like no activity or people sitting/working. Given this, we mark all the different events (what we call outliers), including for example, people walking in the range of the camera or any other different activity. We create a dataset of 10 surveillance videos with different scenarios. Most of these videos have less activity. Given a set $S_1, S_2, \cdots S_k$ of these events marked in the video, define $D(X) = \sum_{i = 1}^k \min(|X \cap S_i|, 1)$. Note this measure is similar to the representative evaluation criteria ($R(X))$ except that it is defined w.r.t the outlier events. Figure~\ref{quantres} (middle left) shows the comparison of the performance of different models on this dataset. As expected, the Diversity measures outperforms the other models consistently.

\paragraph{Extractive Summarization: Importance} To demonstrate the benefit of having the right \emph{importance} or \emph{relevance} terms, we take a set of videos where intuitively the relevance term should matter a lot. Examples include sports videos like Soccer. To demonstrate this, we train a model to predict important events of the video (e.g. the goals, red card). We then define a simple Modular function where the score is the output of the classifier. We then test this out and compare the importance model to other summarization models. The results are shown in Figure~\ref{quantres} (middle right). As we expect, the model with the importance gets the highest scores.

\paragraph{Query Summarization: Diversity} We next look at query based summarization. The goal of query based summarization is to obtain a summary set of frames which satisfy a given query criteria. Figure~\ref{results} (third row) qualitatively shows the results for the query "Sky Scrapers". The Diversity measure is able to obtain a diversity of the different scenes. Even if there is an over-representation of a certain scene in the set of images satisfying the query, the diversity measure tends to pick a diverse set of frames. The representation measure however, tends to focus on the representative frames and can pick more than one image in the summary from scenes which have an over-representation in the query set. We see this Figure~\ref{results}. To quantify this, we define a measure $M(X)$ by dividing the video into a set of clusters of frames $S_1, \cdots, S_k$ where each cluster contains similar frames. These are often a set of continuous frames in the video. We evaluate this on a set of 10 travel videos, and compare the different models. We see that the diversity and representation models tend to perform the best (Figure~\ref{quantres}, bottom left), with the diversity model slightly outperforming the representative models. We also observe that there are generally very few outliers in the case of query based summarization, which is another reason why the diversity model tends to perform well.

\paragraph{Entity Summarization: } Lastly we look at Entity summarization. The goal here is to obtain a summary of the entities (faces, objects, humans) in the video. Figure~\ref{results} (second row) demonstrates the results for Entity summarization of Faces. We see the results for Diversity, Coverage and Representation Models. The diversity model tends to pick up outliers, many of which are false positives (i.e. not faces). The representation model skips all outliers and tends to pick representative faces. To quantitavely evaluate this, we define a representation measure as follows. We remove all the outliers, and cluster the set of entities (objects, faces) into a set of clusters $E_1, \cdots, E_k$ where $E_i$ is a cluster of similar entities. We evaluate this again on a set of 10 videos. Figure~\ref{quantres} (bottom right) shows the results for objects. The results for Faces is similar and in the interest of space, we do not include these. We see that the representation model tends to outperform the other models and skips all the outliers. The diversity model focuses on outliers and hence does not perform as well.

\paragraph{Scalability} Finally, we demonstrate the computational scalability of our framework. Table 2 shows the results of the time taken for Summarization for a two hour video (in seconds) with and without memoization. The groundset size is $|V| = 7200$. We see huge gains from using memoization compared to just computing the gains using the Oracle models of the functions. All our experiments were performed on Intel(R) Xeon(R) CPU E5-2603 v3 @1.6 GHz (Dual CPU) with 32 GB RAM. We used a NVIDIA 1080 GTX 8GB GPU for the Deep Learning. For the two hour video, the preprocessing took around 20 minutes on a single GPU. It would be much faster on multiple GPUs and moreover, this is typically done only once. 

\section{Conclusion}
This paper presents a unified picture of multi-faceted video summarization for extractive, query based and entity based summarization. In each case, we take a closer look at the different summarization models and argue the benefits of these models in different domains. We qualitatively and quantitatively argue this by comparing the results on several domains. Finally, we discuss various implementation tricks to build applications around video and image summarization in production systems.

\bibliographystyle{ieee}
\bibliography{iccv}

\end{document}